\documentclass{article}
\usepackage{spconf,amsmath,graphicx}
\usepackage{float}
\usepackage{hyperref}
\title{SDBERT: S\lowercase{parse}D\lowercase{istil}BERT, \lowercase{a faster and smaller} BERT \lowercase{model}}
\name{Devaraju Vinoda, Pawan Kumar Yadav}
\address{Dept. of CSA, Indian Institute of Science, Bengaluru}
\begin{document}
%\ninept
%
\maketitle
\begin{abstract}
In this work we introduce a new transformer architecture called SparseDistilBERT (SDBERT), which is a combination of sparse attention and knowledge distillantion (KD).
We implemented sparse attention mechanism to reduce quadratic dependency on input length to linear. In addition to reducing computational complexity of the model, we used knowledge distillation (KD).  We were able to reduce the size of BERT model by $60\%$ while retaining $97\%$ performance and it only took $40\%$ of time to train. 

% The abstract should summarize your work with contributions and results.
\end{abstract}
%
% \begin{keywords}
% Sparse attention, Knowledge distillation(KD)
% % one, two, three, four, five
% \end{keywords}
% %
\section{Introduction}
\label{sec:intro}
Transformer\cite{attention} based models have shown to be useful in many NLP tasks. However, a major limitation of transformer based model is its quadratic $O(n^2)$ time and memory complexity (\textit{n} is input sequence length). Hence, it is computationally very expensive to apply transformer based models on long sequences $n>512$. To address these challenges many variant of transformer model have been proposed \cite{bigbird}\cite{etc} which uses sparse attention to scale-up training. Also the transformer based architectures keep getting larger and larger, in 2019 NVIDIA built a model having 8.3 billion parameters. Such bigger models pose challenges under low latency conditions. To address this issue we used distillation\cite{distilbert}\cite{hinton}, a technique to compress a large model called the teacher, into a smaller model called the student. Our main motivation was to focus on reducing computational complexity and simultaneously reduce size of the transformer model.
 
% Empirical studies have shown that for extracting useful information from large amount of data, recommender systems are more effective and reliable than keyword-based search techniques\cite{youtube} \cite{personal}. Scientific papers contain domain-specific specialized information, which makes it hard to search them using only keyword-based approaches.
% Microsoft Academic \cite{microacade} uses hybrid (based  on content and co-citation) technique but it relies more on citations to give recommendations. Our main motivation was to create recommender which does not rely on citations to give recommendations so even less famous but intriguing papers will also get recommended.    

% Discuss briefly about the topic and the related work to your topic published before.

\section{Technical Details}
\label{sec:technical}
% In this section we will discuss two different approaches that we have used to reduce computational complexity. 

% The recommendations are then obtained via cosine similarity measure.
\subsection{SDBERT: SparseDistilBERT}
\label{ssec:use}
We implemented sparse attention\cite{bigbird} instead of full attention and in sparse attention we limit each token to attend to a subset of the other tokens, thereby decreasing the compute time. We used global, sliding \& random tokens instead of attending to all other tokens.
We built a sparse attention based model having 127 million parameters, let us call it teacher model. We used knowledge distillation\cite{tang2019distilling}(teacher-student learning) to reproduce the behavior of the larger teacher model. The smaller model had only 51 million parameters (60\% smaller). Student model has 4 attention heads and 3 hidden layers whereas both were 12 in teacher model. 
For distillation we used the loss function as shown in equation \ref{eq:loss}.
Where $\mathcal{L}_\text{CE}$ is cross entropy loss between true and student predicted labels, $\mathcal{L}_\text{distill}$ is mean-squared-error between teacher logits$(\pmb z^{(B)})$ and student logits$(\pmb z^{(S)})$, and $\alpha$ is weight for the student loss. 

% For understanding Knowledge distillation we have followed paper \cite{lau2016empirical}. We have applied Knowledge distillation on two different models, first model was "attention is All you need" and the dataset that being used is $BBC-News$ Dataset. Second model was "BERT" and the dataset that being used is IMDB dataset. We have used mean-square-error to calculate distill loss  between Teacher soft labels/logits and student soft labels. And final loss is calculated using below formula -    
% \begin{figure}[ht]
%     \centering
%         \includegraphics[width=0.9\linewidth]{formula.jpg}
%         \caption{Relevance score plot. X-axis represents 3 paper each from 3 categories, y-axis represents score (scale of 10).}
%     \label{fig:score}
% \end{figure}

\begin{align}\label{eq:loss}
&\mathcal{L} =\ \alpha\cdot \mathcal{L}_\text{CE} + (1-\alpha)\cdot\mathcal{L}_\text{distill}\\ \nonumber
&\mathcal{L}=-\alpha\sum_i t_i\log y_i^{(S)} - (1-\alpha)||\pmb z^{(B)} - \pmb z^{(S)}||^2_2
\end{align}

\begin{figure}
    \centering
        \includegraphics[width=0.9\linewidth]{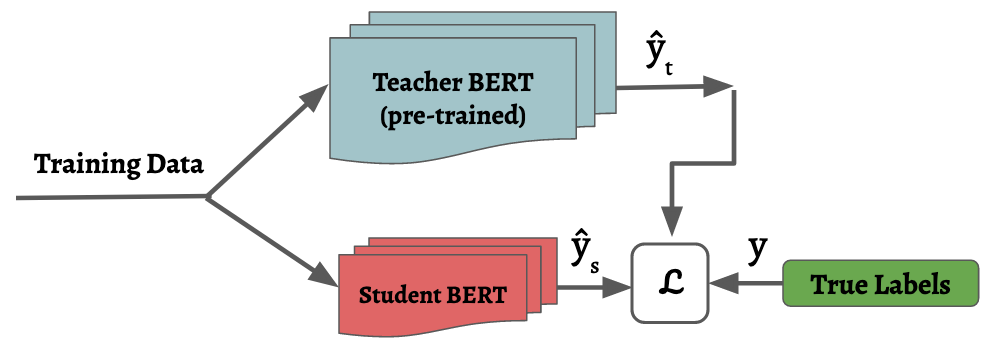}
        \caption{SDBERT architecture, first we used sparse attention to train a teacher BERT model. Then we reduced the number of attention heads and hidden layers to get a student BERT model and applied knowledge distillation.}
    \label{fig:score}
\end{figure}
\vspace{-8mm}%Put here to reduce too much white space after your table 
% USE \cite{use_enc} encodes text into 512 dimensional vector and it is a transformer based sentence encoding model.
\section{Results}
To evaluate our approach we used IMDb dataset (containing positive/negative sentiment) and results are shown in table \ref{tab:res}, accuracy along with training time.
% With combination of sparse attention and KD we were able to achieve an accuracy of $87.9\%$ instead of $70.3\%$ without KD. 
\label{sec:results}
\begin{center}
\begin{table}[H]
\begin{tabular}{ lc c c }
 \hline
 \textbf{Model} & \textbf{Accuracy} & \textbf{Time \textit{(min)}} \\ 
 \hline
 Full Attention & 93.3 & 132.4 \\  
 Sparse Attention & 90.6 & 115.5 \\
 Sparse Attention with KD & 87.9 & 46.74\\
 Sparse Attention without KD & 70.3 & 16.77\\
 \hline
\end{tabular}
\caption{Test results on IMDb dataset.}
\label{tab:res}
\end{table}
\end{center}
\vspace{-13mm}%Put here to reduce too much white space after your table 
\section{Contributions}
\label{sec:contributions}
\vspace{-3mm}%Put here to reduce too much white space after your table 
Existing works focuses on either reducing compute(sparse attention) or compressing the models (KD). So far no one has tried to use KD and sparse attention together. In this project we combined both ideas and implemented in a single model. We have written all the code on our own by referring on relevant library documentations. Code can be found at this GitHub \href{https://github.com/devarajuvinoda/SparseDistilBERT-SDBERT-}{\textit{repo}}
% \vspace{10pt}
% Describe the novel aspect of your work which can be re-implementation of the codes from scratch or  addition of your own component in the pre-existing work

% Below is an example of how to insert images. Delete the ``\vspace'' line,
% uncomment the preceding line ``\centerline...'' and replace ``imageX.ps''
% with a suitable PostScript file name.
% -------------------------------------------------------------------------
% \begin{figure}[htb]

% \begin{minipage}[b]{1.0\linewidth}
%   \centering
%   \centerline{\includegraphics[width=8.5cm]{image1}}
% %  \vspace{2.0cm}
%   \centerline{(a) Result 1}\medskip
% \end{minipage}
% %
% \begin{minipage}[b]{.48\linewidth}
%   \centering
%   \centerline{\includegraphics[width=4.0cm]{image3}}
% %  \vspace{1.5cm}
%   \centerline{(b) Results 3}\medskip
% \end{minipage}
% \hfill
% \begin{minipage}[b]{0.48\linewidth}
%   \centering
%   \centerline{\includegraphics[width=4.0cm]{image4}}
% %  \vspace{1.5cm}
%   \centerline{(c) Result 4}\medskip
% \end{minipage}
% %
% \caption{Example of placing a figure with experimental results.}
% \label{fig:res}
% %
% \end{figure}

%\vfill
%\pagebreak

\pagebreak
\section{Resources}
\label{sec:resources}
We used subsets of IMDb data, that is available for access to customers for personal and non-commercial use. It has total 50k data examples containing movie review with sentiment (either positive or negative).

\bibliographystyle{IEEEbib}
\bibliography{strings,refs}
\end{document}